\documentclass[11pt]{article}

\usepackage{arxiv}
\usepackage[numbers]{natbib}

\pdfoutput=1
\usepackage{times}
\usepackage{latexsym}

\usepackage[T1]{fontenc}

\usepackage[utf8]{inputenc}

\usepackage{microtype}

\usepackage{inconsolata}

\usepackage{amsmath}
\usepackage{multirow}           
\usepackage{cleveref}
\usepackage{adjustbox}
\usepackage{booktabs}
\usepackage{cleveref}

\newcommand{\eg}{\textit{e.g\@.}}

\newcommand{\etal}{\textit{et~al\@.}}

\usepackage{tikz}

\title{Dynamic Stashing Quantization for Efficient Transformer Training}

\author{Guo Yang \\
  University of Cambridge \\
  \texttt{gy261@cam.ac.uk} \\\And
  Daniel Lo \\
  Microsoft \\
  \texttt{dlo@microsoft.com} \\\AND
  Robert Mullins \\
  University of Cambridge \\
  \texttt{robert.mullins@cl.cam.ac.uk} \\\And
  Yiren Zhao \\
  Imperial College London \\
  \texttt{a.zhao@imperial.ac.uk}
  }

\begin{document}
\maketitle

\begin{abstract}
    Large Language Models (LLMs) have demonstrated impressive performance on a range of Natural Language Processing (NLP) tasks. Unfortunately, the immense amount of computations and memory accesses required for LLM training makes them prohibitively expensive in terms of hardware cost, and thus challenging to deploy in use cases such as on-device learning.
    
    In this paper, motivated by the observation that LLM training is memory-bound, we propose a novel dynamic quantization strategy, termed \emph{Dynamic Stashing Quantization} (DSQ), that puts a special focus on reducing the memory operations, but also enjoys the other benefits of low precision training, such as the reduced arithmetic cost. We conduct a thorough study on two translation tasks (trained-from-scratch) and three classification tasks (fine-tuning). DSQ reduces the amount of arithmetic operations by $20.95\times$ and the number of DRAM operations by $2.55\times$ on IWSLT17 compared to the standard 16-bit fixed-point, which is widely used in on-device learning.
\end{abstract}

\section{Introduction}

Large Language Models (LLMs) based on the Transformer architectures \cite{vaswani2017attention}
are currently seen as the foundation models \cite{bommasani2021opportunities}. 
The \emph{pre-train and then fine-tune} paradigm has shown promising results for a variety of Natural Language Processing (NLP) tasks \cite{liu2019roberta,raffel2020exploring,brown2020language}.
However, the training of LLM is both computationally and memory intensive, posing a significant challenge for their deployment.

In the hardware world, 
the \emph{Roofline model } demonstrates that there is an optimal balance of processor and memory performance. The metric used to assess performance is referred to as the \emph{operational intensity}, which is calculated as the ratio of arithmetic intensity to memory bandwidth:

\begin{figure}[!ht]
    \centering
    \includegraphics[scale=0.55]{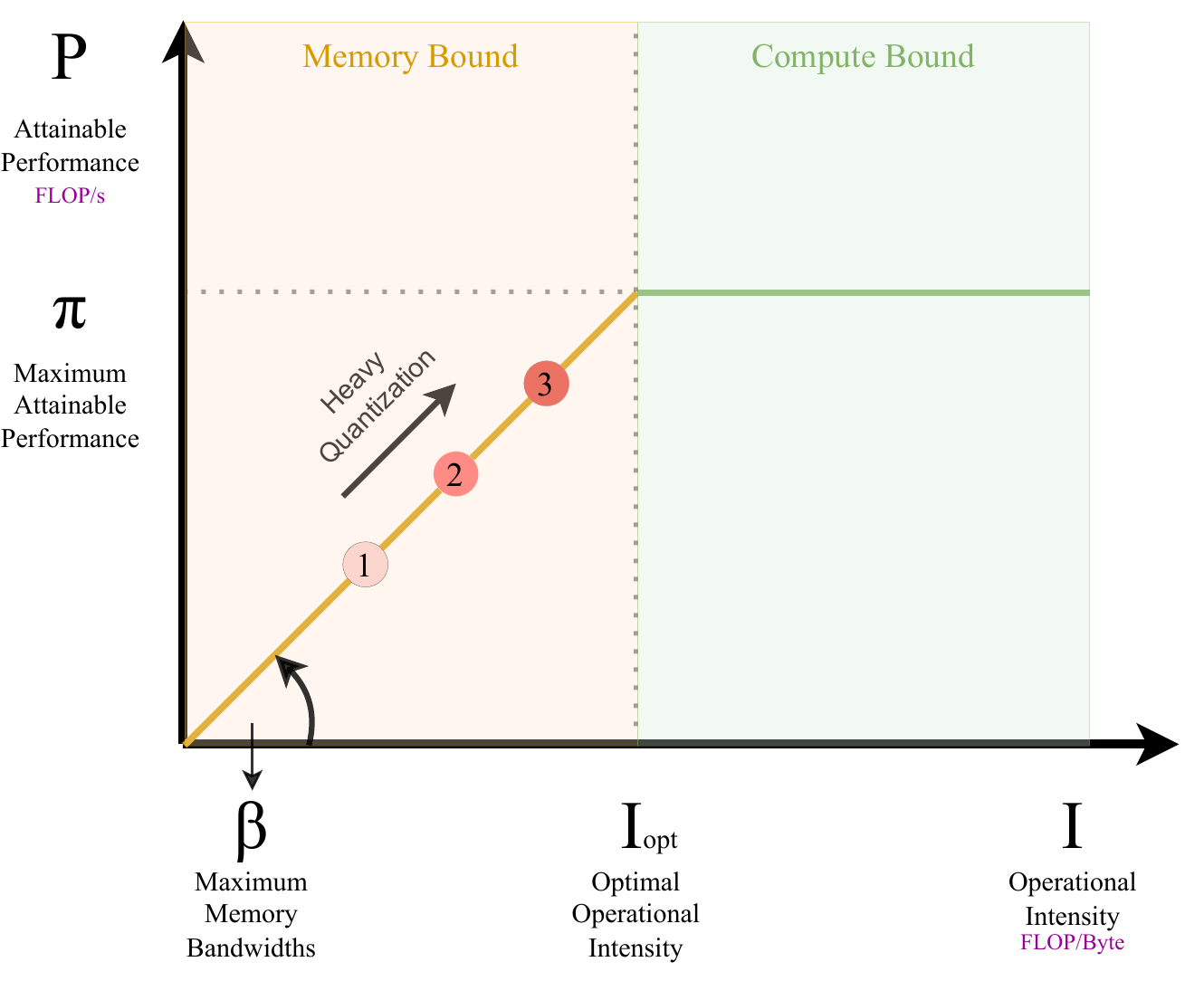}
    \caption{The Roofline model with operational intensity ($I$) and attainable performance ($P$). 1 is non-quantized, 2 is a standard quantization and 3 is DSQ.}
    \label{fig:roofline}
\end{figure}

\begin{equation*}
	\text{Operational Intensity} = \frac{\text{Number of Operations}}{\text{DRAM traffic}}
\label{eq:operational_intensity}
\end{equation*}

The Roofline model has enabled us to identify the sweet spot ($I_{opt}$) for a processor to reach its peak arithmetic performance \cite{williams2009roofline,ding2022instruction}.
As illustrated in \Cref{fig:roofline}, as operational intensity ($I$) increases, the maximum attainable performance rises at a linear rate initially before reaching a constant value. The region to the left of the turning point is limited by the available memory bandwidth; the region to the right is constrained by the processor's arithmetic computing capability. Training Transformer models, as shown by Ivanov \etal~\cite{ivanov2021data}, is memory-bound, which means it sits at the left quadrant in the Roofline model ($I < I_{opt}$). Consequently, the performance of LLM training on modern hardware is significantly hindered by the inadequate bandwidth, as the amount of data movements to and from DRAM is the major performance bottleneck. 

For this reason, researchers have sought to accelerate the training process of Transformers through \emph{quantization}. 
Prior work has looked into the effect of quantization on Transformer models, a majority of which focus on the forward pass of model inference with fixed weights \cite{zhang2020ternarybert,bai2020binarybert,tao2022compression}. A number of studies have also investigated low-precision training for Transformers \cite{sun2019hybrid,sun2020ultra}. 
Although works have demonstrated the effectiveness of quantization, they typically assume a single precision level, either per neural network layer or per network, which over-simplifies the hardware target. When viewed from a Roofline model perspective, \emph{existing quantization methods attempt to optimize both compute complexity and memory bandwidth requirement, and then fail to recognize that the workload is heavily memory-bound.}



Motivated by this observation, we propose a novel quantization strategy for LLM training named \emph{Dynamic Stashing Quantization} (DSQ).
We identify the most memory-intensive part of LLM training -- the communication between the forward and backward passes, and define \emph{stashing} as the process of storing intermediate results in a memory buffer (in a normal case, DRAM) for later use.
The proposed quantization places an emphasis on this communication, and \emph{dynamically quantize the intermediate results between forward and backward passes} for a significant reduction of the DRAM traffic. As illustrated in \Cref{fig:roofline}, this reduction of DRAM bandwidth helps DSQ to move closer to the optimal operational intensity.
We have the following contributions:
\begin{itemize}
	\item We propose \emph{Dynamic Stashing Quantization (DSQ)} for LLM training. DSQ not only quantizes operations for the entire training process, but also employs a \emph{more aggressive quantization} for intermediate results between the forward and backward passes to drastically minimize DRAM traffic.
	\item DSQ follows a time-adaptive principle for stashing, which involves starting with lower precision at the beginning of the training process and gradually increasing the precision as it progresses. DSQ has been demonstrated to provide a higher performance compared to its fixed-precision counterpart.
	\item We evaluate the proposed strategy on a variety of tasks and setups, including training from scratch and fine-tuning.  DSQ achieves up to a $2.55\times$ increase in arithmetic performance and a $20.95\times$ reduction in DRAM requirement compared to 16-bit fixed-point training.
\end{itemize}

\section{Related Work}


Quantization has been studied in detail for inference of neural network models. These include using uniform \cite{zafrir2019q8bert,bhandare2019efficient} and non-uniform \cite{sun2019hybrid,darvish2020pushing} quantization methods. Specifically, uniform quantization methods such as fixed-point \cite{zafrir2019q8bert,bhandare2019efficient,lin2020towards}, ternary \cite{zhang2020ternarybert}, or even binary \cite{bai2020binarybert} number formats have been applied to inference of Transformer models. In this work, we focus on quantization for LLM training which introduces new challenges such as the large dynamic range needed during the backward pass for lossless training \cite{sun2019hybrid} where non-uniform quantization methods have seen more success.

Training LLM models is approximately $3 \times$ more expensive than running inference for the same model. 
Thus, quantizing all operations during training has been an area of active research
\cite{sun2019hybrid,sun2020ultra,yang2019swalp,fu2020fractrain,fox2020block,kalamkar2019study}. 
Most of these methods use non-uniform quantization to handle larger dynamic range needed for gradient updates \cite{kalamkar2019study}.
Floating-point arithmetic and its variants have become a popular method for low-precision training (\eg~fewer than 8 bits). Mini-floats with extremely small exponents (\eg~1 bit or 2 bits) have been demonstrated to be effective in small language models, such as LSTMs \cite{sun2019hybrid,sun2020ultra}.
Block floating-point or block mini-floats, where an exponent is shared between a set of values, has become popular in quantized training \cite{yang2019swalp,drumond2018training,fox2020block} as it allows for a large dynamic range while approximating the cost of integer formats for multiplication.
Specifically, Draumond \etal~utilized block floating-point with roughly 24 bits to perform lossless training on vision tasks \cite{drumond2018training}. Fox \etal~demonstrated that 8-bit training is possible with an around $0.5$ BLEU score degradation on machine translation \cite{fox2020block}. Our work extends these formats to Large Language Models, includes quantization of stashed weights, and introduces a dynamic aspect to further reduce the required bit widths. The idea of \emph{stashing} has also been explored before by Jain \etal, although they only focused on applying lossless encoding methods on single precision numbers (Float16) \cite{jain2018gist}. However, in this paper we show a more aggressive stashing techniques (\eg~on average less than 4 bits per number) that is time-adaptive for LLM training.
Fractrain \cite{fu2020fractrain}, to our knowledge, is the only work that applied the idea of dynamic quantization on standard training, but was primarily focusing on visition tasks. Our work extends dynamic quantization to encompass stashed values and evaluates these effects on LLMs. 
Prior research on distributed training has looked at reducing the communication cost \cite{alistarh2017qsgd,honig2022dadaquant}, where Honig \etal~ also investigate how a time-adaptive quantization would help federated systems to learn. These works focused on device-to-device traffic while our work focuses on reducing DRAM traffic.


\section{Method}
\begin{figure*}[h]
    \centering
    \includegraphics[scale=0.35]{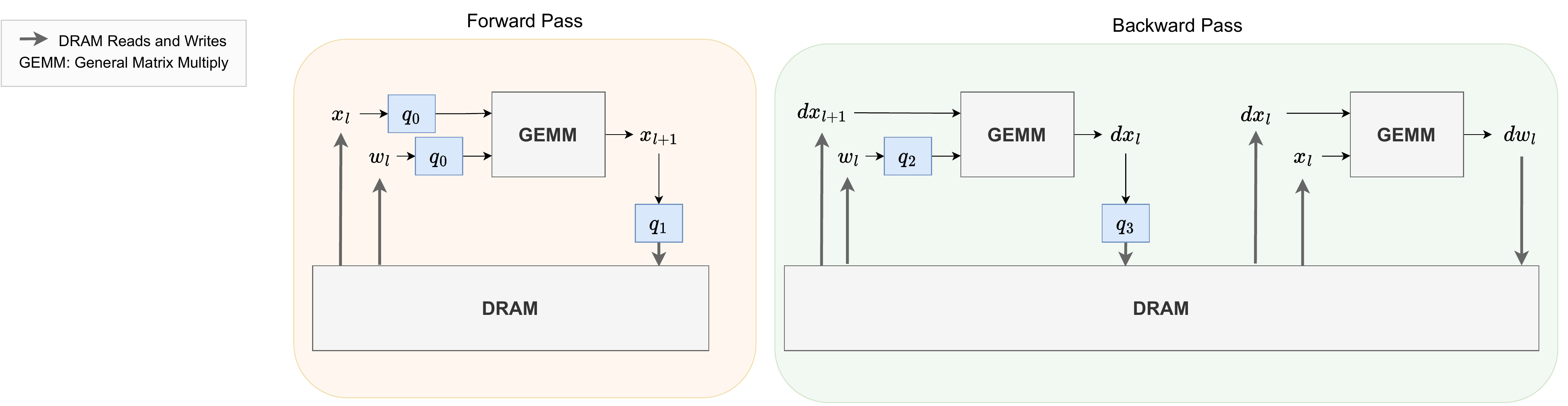}
    \caption{An illustration of the DSQ flow for a single linear layer. 
    The training is viewed as a combination of a forward pass and a backward pass.
    $q_0$, $q_1$, $q_2$ and $q_3$ define where the tensors are quantized, we use $[q_0, q_1, q_2, q_3]$ to describe the DSQ configuration. DSQ ensures all GEMM inputs are quantized. Notice for the second and third GEMMs, $dx_{l+1}$, $x_l$ and $dx_l$ are the quantized version fetched from the DRAM, the fact that these values are heavily quantized helps us to save DRAM bandwidth.}
    \label{fig:method}
\end{figure*}

\Cref{fig:method} provides a high-level illustration of the DSQ flow. We consider the inputs $x_l$ of a neural network layer with parameters $w_l$, and the output of the layer is $x_{l+1}$. In the backward pass, we consider the partial derivatives $dx_{l}$ of the input and also the gradient of the weights $dw_l$. Naturally, a single training step requires three GEMMs as illustrated in \Cref{fig:method}. We illustrate four quantization opportunities in this training step and their effects:
\begin{itemize}
    \item $q_0$: mainly affects the arithmetic density of forward pass, notice it is possible for $x_l$ and $w_l$ to use different precisions, but this optimization is not the focus of our work.
    \item $q_1$: affects the DRAM memory bandwidth, one key point in our work is that we show $q_1$ can be different from $q_0$ and in fact can be a very aggressive, dynamic quantization.
    \item $q_2$: affects mainly the computation complexity of the first GEMM in the backward pass.
    \item $q_3$: affects the DRAM bandwidth and also the computation complexity of the second GEMM in the backward pass.
\end{itemize}

In our knowledge, we are the first to systematically illustrate the potential effects, both on compute and off-chip memory bandwidth, of various quantization opportunities within a standard training pass. The two GEMMs in the backward pass can be potentially fused (\eg~pipelined), and in that case $dx_l$ does not have to be written to and then read from the DRAM. In our cost model estimation, we use a conservative strategy and assume this tensor is always flushed to DRAM. In DSQ, we use Block Floating Point as the quantizer for $q_0$, $q_1$, $q_2$ and $q_3$, since this quantizer is shown superior to fixed-point quantization \cite{darvish2020pushing}. 
We also use a time-adaptive quantization strategy, this means the quantization uses a different quantization level $q_i^t$ for each round $t$ of the training. We design DSQ to monotonically increase $q_i^t$ as a function of $t$ and use the validation loss to inform this increase. This monotonic increase strategy has been proven more effective than other complex scheduling methods in \citeauthor{honig2022dadaquant}. Through extensive tuning and experimentation, we also notice that it is important to keep $q_3 \geq 16$ through the entire training process, and \Cref{sec:appendix:q16} studies the effect of different quantization levels for $q_3$.

\section{Evaluation}
\begin{table*}[!ht]
	\centering
	 \caption{The performance of Machine Translation trained with a 6-layer Transformer architecture \cite{vaswani2017attention}, the model is assessed using numbers reported as percentages. $\Delta$ shows the performance difference compared to the floating-point 32-bit baseline.}
	\adjustbox{scale=0.70}{
		\begin{tabular}{llcccc}
			\toprule
			Dataset and Model                         
			& Method                   
			& Precision Setup      
			& Acc / BLEU ($\Delta$)
			& Arith Ops ($\downarrow$) 
			& DRAM R/W ($\downarrow$) \\
			\midrule
			\multirow{7}{4cm}{IWSLT2017 DE-EN Transformer (6-layer)} 
                & Floating-point        
			& {[}32, 32, 32, 32{]}    &   $35.22$        &    -       &    -     \\
			& Fixed-point           
			& {[}32, 32, 32, 32{]} 
			& $34.47$ $(-0.75)$         
			&  $1.00\times$         
			&  $1.00\times$       \\
			& Fixed-point              
			& {[}16, 16, 16, 16{]} 
			&  $32.59$ $(-2.63)$         
			&  $0.25\times$         
			&  $0.50\times$       \\
			& Block FP                 
			& {[}32, 32, 32, 32{]} 
			& $34.56$ $(-0.66)$          &  $0.56\times$         &   $1.13\times$      \\
			& Block FP                 
			& {[}16, 16, 16, 16{]}     &  $34.30$ $(-0.92)$         &   $0.18\times$        &    $0.63\times$     \\
			& Stashing (Fixed)         
			& {[}16, 4, 4, 16{]}     &  $25.50$ $(-9.72)$        &   $0.13\times$        &    $0.31\times$     \\
			& Stashing (BFP)           
			& {[}16, 4, 4, 16{]}     &  $34.78$ $(-0.44)$         &   $0.10\times$        &     $0.45\times$    \\
			& DSQ (BFP)   &  $-$                    &$34.81$ $(-0.41)$           &   $0.012\times$        &     $0.20\times$    \\
    \midrule
			\multirow{8}{4cm}{GLUE MNLI \\ RoBERTa-base} 
                & Floating-point         
			&  {[}32, 32, 32, 32{]}   &   $87.6$        &    -       &    -     \\
			& Fixed-point           
			& {[}32, 32, 32, 32{]} 
			&  $87.9$ $(+0.3)$        
			&  $1.00\times$          
			&  $1.00\times$       \\
			& Fixed-point              
			& {[}16, 16, 16, 16{]} 
			&  $87.9$ $(+0.3)$    
			&  $0.25\times$         
			&  $0.50\times$       \\
			& Block FP                 
			& {[}32, 32, 32, 32{]} 
			&  $87.8$ $(+0.2)$       &  $0.56\times$         &  $1.13\times$       \\
			& Block FP                 
			& {[}16, 16, 16, 16{]}     & $87.8$ $(+0.2)$         &  $0.18\times$         &  $0.63\times$       \\
			& Stashing (Fixed)         
			& {[}16, 4, 4, 16{]}     & $82.8$ $(-4.8)$         &   $0.13\times$        &   $0.32\times$      \\
			& Stashing (BFP)           
			& {[}16, 4, 4, 16{]}     & $87.8$ $(+0.2)$          &   $0.10\times$        &    $0.45\times$     \\
            & DSQ (BFP)           
			& $-$     & $87.8$ $(+0.2)$          &   $0.043\times$        &  $0.26\times$       \\
			
    \midrule
			\multirow{8}{4cm}{GLUE QNLI \\ RoBERTa-base} 
                & Floating-point         
			& {[}32, 32, 32, 32{]}    &   $92.8$        &  -         &       -  \\
			& Fixed-point           
			& {[}32, 32, 32, 32{]} 
			&   $92.6$ $(-0.2)$       
			&  $1.00\times$         
			&  $1.00\times$       \\
			& Fixed-point              
			& {[}16, 16, 16, 16{]} 
			&  $92.6$ $(-0.2)$         
			&  $0.25\times$         
			&  $0.50\times$       \\
			& Block FP                 
			& {[}32, 32, 32, 32{]} 
			&   $92.7$ $(-0.1)$        &   $0.56\times$        &     $1.13\times$    \\
			& Block FP                 
			& {[}16, 16, 16, 16{]}     & $92.5$ $(-0.3)$          &   $0.18\times$        &  $0.63\times$       \\
			& Stashing (Fixed)         
			& {[}16, 4, 4, 16{]}     & $89.5$ $(-3.3
   )$           &  $0.13\times$         &   $0.32\times$      \\
			& Stashing (BFP)           
			& {[}16, 4, 4, 16{]}     &  $92.6$ $(-0.2)$          &  $0.10\times$         &  $0.45\times$       \\
            & DSQ (BFP)           
			& $-$    &  $92.7$ $(-0.1)$          &  $0.043\times$         &  $0.26\times$       \\
			
			\bottomrule
			\end{tabular}
	 }

	 \label{tab:all}
	\end{table*}

We evaluate the effectiveness of DSQ on two different translation tasks, WMT14 EN-DE \cite{bojar2014findings} (in \Cref{sec:appendix:wmt}) and IWSLT17 EN-DE \cite{cettolo2017overview}, and two tasks from the GLUE benchmark \cite{wang2018glue}, the details of these datasets are in \Cref{sec:appendix:datasets}.
We used the Adam optimizer and the details for all the learning rate and batch size selections are in \Cref{sec:appendix:params}. For the translation tasks, we use a classic 6-layer transformer model \cite{vaswani2017attention} and the RoBERTa-base model \cite{liu2019roberta} for the GLUE tasks. All tasks are executed on systems that have 2 AMD EPYC 7763 64-Core Processors 1.8GHz (128 cores in total),  and 4 NVIDIA A100-SXM-80GB GPUs, with 1000 GiB RAM.
We are interested in understanding the costs of arithmetic operations, as well as the number of memory reads and writes. To this end, we have built a hardware performance modeling framework to estimate the training cost. Our cost model is similar to \citeauthor{sun2020ultra} and \citeauthor{samajdar2018scale}, but our numbers are derived from a production hardware system, taking the numbers reported in \citeauthor{darvish2020pushing}, to provide a higher-fidelity estimation.






\Cref{tab:all} presents the results of our study comparing different quantization strategies. We compare not only popular low-latency training baselines such as 32-bit floating-point, 32 and 16-bit fixed-point; but also Block floating-point (BFP) \cite{darvish2020pushing,fox2020block} with different precisions. For all BFP implementations considered in this paper, we keep the exponent bitwidth to be $8$ and the bounding-box size to be $16$ following \citeauthor{darvish2020pushing}. In addition, we compare static stashing strategies that are based on either fixed-point (Fixed) or BFP. In \Cref{tab:all}, we use the hardware cost of fixed-point 32-bit computation as $1 \times$ since this is a stronger baseline.
The results in \Cref{tab:all} demonstrate that DSQ has a comparable accuracy and BLEU score compared to 32-bit fixed-point for training while having a $20.95\times$ reduction in arithmetic complexity and a $2.55\times$ decrease in DRAM Read and Writes on IWSLT14 DE-EN. DSQ also show very competitive accuracy on fine-tuning RoBERta on GLUE while having a much smaller hardware utilization.

\section{Conclusion}
\vspace{-10pt}
In this paper, we propose Dynamic Stashing Quantization (DSQ) for LLM training. This new quantization strategy applies a more aggressive quantization for intermediate results between the forward and backward passes generated during training, thereby reducing DRAM traffic. Specifically, our approach uses a low precision at the beginning of training, and then gradually increases the precision level, to reduce the effect of round-off errors introduced by quantization. We demonstrate the effectiveness of DSQ by showing how it can reduce both the computation cost and DRAM bandwidth requirement on machine translation and LLM fine-tuning tasks.
\newpage

\bibliography{custom}
\bibliographystyle{plainnat}

\newpage

\appendix

\section{Datasets}
\label{sec:appendix:datasets}
\begin{table*}[!ht]
\centering
 \caption{Details for each dataset, including the number of classes, a description and the source.}
 \label{tab:dataset_setup}
\adjustbox{scale=0.8}{
	\begin{tabular}{c | c | p{10cm} }
	\toprule
	Name & \# Class & Description \\
        \midrule
	\multirow{2}{*}{WMT14 EN-DE} 
        & \multirow{2}{*}{-}
        & A text translation task on English-German sentence pairs from The The Stanford Natural Language Processing Group.\\
        
        \midrule
	\multirow{2}{*}{IWSLT2017 DE-EN} 
        & \multirow{2}{*}{-}
        & A text translation task on German-English sentence pairs from The International Conference on Spoken Language Translation.\\
        
	\midrule
	\multirow{3}{*}{QNLI} 
        & \multirow{3}{*}{2}
        & A binary textual entailment task on question-answer pairs from the Stanford Question Answering database. The objective is to determine whether a pair is an entailment or not. \\

        \midrule
	\multirow{5}{*}{MNLI} 
        & \multirow{5}{*}{3}
        & A multi-class (i.e., entailment, neutral, contradiction) textual entailment task on premise-hypothesis pairs from the Multi-genre Natural Language Inference corpus. Matched version only preserves pairs within the same genre (e.g., government report, science fiction, speech). \\

        \bottomrule
        \end{tabular}
 }

\end{table*}
Four datasets are used: translation WMT14 EN-DE and IWSLT2017 EN-DE for machine translation tasks, QNLI and MNLI for textual entailment tasks. \Cref{tab:dataset_setup} presents details for the datasets.

\section{Hyperparameters}
\label{sec:appendix:params}
\begin{table*}[!ht]
\centering
 \caption{Details of the optimal hyper-parameters including batch size, learning rate and weight decay values for each set of experiments with the same dataset and prompting model.}
 \label{tab:hyper_param}
\adjustbox{scale=.8}{
	\begin{tabular}{c | c c c c}
	\toprule
	Dataset & Batch size & Max tokens & Learning rate & Weight decay    \\
	\midrule
        WMT14 EN-DE
        & -
        & 4096
        & 5e-4
        & 0.0 \\  
        IWSLT2017 DE-EN
        & -
        & 4096
        & 5e-4 
        & 1e-4 \\  
        QNLI
        &  32
        & 4400
        & 1e-5
        & 0.1 \\
        MNLI
        &  32
        & 4400
        & 1e-5
        & 0.1 \\

        \bottomrule
        \end{tabular}
 }

\end{table*}
\begin{table*}[!ht]
	\centering
	 \caption{Tests on stashing precision setup. The models are trained on IWSLT14 DE-EN. $\Delta$ shows the performance difference compared to the floating-point 32-bit baseline.}
  \label{tab:iwslt:different}
	\adjustbox{scale=0.8}{
		\begin{tabular}{llcccc}
			\toprule
			Dataset and Model                         
			& Method                   
			& Precision Setup      
			& Acc / BLEU ($\Delta$)
 \\
			\midrule
			\multirow{7}{4cm}{IWSLT14 DE-EN Transformer (6-layer)} 
			 & Stashing (BFP)           
			 & {[}2, 2, 2, 16{]}     &   $17.45$ $(-17.77)$         \\
              & Stashing (BFP)           
			 & {[}4, 2, 2, 16{]}     &   $33.51$ $(-1.71)$         \\
              & Stashing (BFP)           
			 & {[}4, 4, 4, 16{]}     &   $34.47$ $(-0.75)$         \\
              & Stashing (BFP)           
			 & {[}8, 4, 4, 16{]}     &   $34.47$ $(-0.75)$         \\
              & Stashing (BFP)           
			 & {[}8, 8, 8, 16{]}     &   $34.65$ $(-0.57)$         \\
              & Stashing (BFP)           
			 & {[}16, 4, 4, 16{]}     &   $34.78$ $(-0.44)$         \\
              & Stashing (BFP)           
			 & {[}16, 8, 8, 16{]}     &   $34.47$ $(-0.75)$         \\

			\bottomrule
			\end{tabular}
	 }
	\end{table*}
The training hyperparamters, such as learning rates, are picked following standard benchmarks and open implementaitons \cite{liu2019roberta,vaswani2017attention}. We summarize them in \Cref{tab:hyper_param} for repeatability. We use the Adam optimizer with $\beta_1 = 0.9$, $\beta_2 = 0.98$ for both training and finetuning models. The learning rate schedule is Inverse Square Root for training models, and Polynomial Decay for finetuning models. Dropout with rates of $P_{IWSLT}=0.3$ and $P_{WMT}=0.2$, label smoothing with value $\epsilon=0.1$ are applied to train models.

DSQ precision configurations are decided through experimentation on the IWSLT dataset and then the same setup is used for all other detests. The idea is that after observing several epochs of unchanged or increasing validation loss, the model adapts to a less aggressive precision setup. \Cref{tab:iwslt:different} shows a collectin of tuning runs we had, we found that heavily quantized models still work at the start of training stage, and [16, 4, 4, 16] quantized BFP model works as well as less aggressive ones. This indicates that DSQ should start with heavily aggressive precision setup (we pick [2, 2, 2, 16] for IWSLT14 DE-EN), and jump to [16, 4, 4, 16] when needed during training process.

\section{The effect of $q_3$}
\label{sec:appendix:q16}
\begin{table*}[!ht]
	\centering
	 \caption{Tests on gradient output precision setup. The models are trained on IWSLT14 DE-EN.}
  \label{tab:q16_iwslt}
	\adjustbox{scale=0.8}{
		\begin{tabular}{llcccc}
			\toprule
			Dataset and Model                         
			& Method                   
			& Precision Setup      
			& Acc / BLEU ($\Delta$)
 \\
			\midrule
			\multirow{3}{4cm}{IWSLT14 DE-EN Transformer (6-layer)} 
			 & Stashing (Fixed)           
			 & {[}8, 8, 8, 32{]}     &   $34.08$        \\
              & Stashing (Fixed)           
			 & {[}8, 8, 8, 16{]}     &   $31.94$         \\
              & Stashing (Fixed)           
			 & {[}8, 8, 8, 8{]}     &   Failed      \\

			\bottomrule
			\end{tabular}
	 }
	\end{table*}
The gradient output ($dx_{l}$) plays an important role in the performance of fixed-point quantization. Notice in table \Cref{tab:q16_iwslt}, gradient output quantized to 8 bits leads to training failure for fixed-point quantization. In order to focus on the idea of stashing, we apply 16 bits quantization of gradient output for all our stashing precision setups.

\section{Additional results on WMT14}
\label{sec:appendix:wmt}
\begin{table*}[!ht]
	\centering
	 \caption{The performance of Machine Translation trained on WMT14 EN-DE with a 6-layer Transformer architecture \cite{vaswani2017attention}, the model is assessed using numbers reported as percentages. $\Delta$ shows the performance difference compared to the floating-point 32-bit baseline.}
	 \label{tab:wmt}
	\adjustbox{scale=0.8}{
		\begin{tabular}{llcccc}
			\toprule
			Dataset and Model                         
			& Method                   
			& Precision Setup      
			& Acc / BLEU ($\Delta$)
			& Arith Ops 
			& DRAM R/W \\
			\midrule
			\multirow{7}{4cm}{WMT14 EN-DE Transformer (6-layer)} 
                 & Floating-point         
			 & {[}32, 32, 32, 32{]}   &   $25.79$        &  -         & -        \\
			 & Fixed-point           
			 & {[}32, 32, 32, 32{]} 
			 & $25.41$ $(-0.38)$         
			 & $1.00\times$          
			 & $1.00\times$        \\
			 & Fixed-point              
			 & {[}16, 16, 16, 16{]} 
			 & $23.40$ $(-2.39)$         
			 & $0.25\times$          
			 & $0.50\times$         \\
			 & Block FP                 
			 & {[}32, 32, 32, 32{]} 
			 & $25.76$ $(-0.03)$         & $0.56\times$           &  $1.13\times$        \\
			 & Block FP                 
			 & {[}16, 16, 16, 16{]}     & $25.61$ $(-0.18)$          &  $0.18\times$          &  $0.63\times$        \\
			 & Stashing (Fixed)         
			 & {[}16, 4, 4, 16{]}     &   $21.86$ $(-3.93)$        & $0.13\times$           &   $0.31\times$       \\
			 & Stashing (BFP)           
			 & {[}16, 4, 4, 16{]}     &   $25.24$ $(-0.55)$        &  $0.10\times$          &   $0.20\times$       \\		
			\bottomrule
			\end{tabular}
	 }

	\end{table*}
We also train the model on WMT14 EN-DE dataset, the BLEU scores we gain are relatively low compared to the $27.3$ BLEU score achieved by \citeauthor{vaswani2017attention} because we only trained the models for $15$ epochs. \Cref{tab:wmt} presents the results.



\end{document}